# Spatiotemporal Calibration for Laser Vision Sensor in Hand-eye System Based on Straight-line Constraint

Peiwen Yang, Mingquan Jiang, Xinyue Shen and Heping Zhang

*Abstract*—**Laser vision sensors (LVS) are critical perception modules for industrial robots, facilitating real-time acquisition of workpiece geometric data in welding applications. However, the camera communication delay will lead to a temporal desynchronization between captured images and the robot motions. Additionally, hand-eye extrinsic parameters may vary during prolonged measurement. To address these issues, we introduce a measurement model of LVS considering the effect of the camera's time-offset and propose a teaching-free spatiotemporal calibration method utilizing line constraints. This method involves a robot equipped with an LVS repeatedly scanning straight-line fillet welds using S-shaped trajectories. Regardless of the robot's orientation changes, all measured welding positions are constrained to a straight-line, represented by Plucker coordinates. Moreover, a nonlinear optimization model based on straight-line constraints is established. Subsequently, the Levenberg-Marquardt algorithm (LMA) is employed to optimize parameters, including time-offset, hand-eye extrinsic parameters, and straight-line parameters. The feasibility and accuracy of the proposed approach are quantitatively validated through experiments on curved weld scanning. We open-sourced the code, dataset, and simulation report at https://github.com/RoboticsPolyu/LVS_ST_CALIB.**

## I. Introduction

Machine vision technology endows robotics and automation systems with advanced perception capabilities, playing a pivotal role in intelligent robotic measurement and manufacturing [1-3], such as surface reconstruction [4-6], welding [7-9], and grinding [10]. The two key factors in laser vision sensor (LVS) calibration in automated robotic system are time-offset and hand-eye calibration [11-13]. The time-offset calibration facilitates the temporal data association between robot's motions and captured images by the LVS. In addition, hand-eye calibration determines the coordinate transformation between the LVS and the industrial robot. In this study, a spatiotemporal calibration method for LVS in the hand-eye system is proposed.

Manuscript received: March 31, 2025; Revised August 3, 2025; Accepted: October 22, 2025.

This paper was recommended for publication by Editor Lucia Pallottino upon evaluation of the Associate Editor and Reviewers' comments.

This research was funded by the Natural Science Foundation of Guangxi Province under the project titled 'Research on the 3D Coverage Mechanism of Intelligent Sensor Networks for Water Quality Monitoring in River Basins' (Grant No. 2025GXNSFAA0691009). Additional funding was provided by the Guangxi University Young and Middle-Aged Teachers' Research Basic Ability Enhancement Project, under the title 'Research on Knowledge Graph-Based Fault Diagnosis for CNC Machine Tools Using Deep Learning' (Grant No. 2025KY0809).

Peiwen Yang is with the Department of Aeronautical and Aviation Engineering, Hong Kong Polytechnic University, China. Mingquan Jiang is with the School of Artificial Intelligence, Hezhou University, China. Xinyue Shen is with the School of Mechatronic Engineering and Automation, Shanghai University, China. Heping Zhang is with the School of Mathematics and Statistics, Central China Normal University, China. (e-mail: peiwen1.yang@connect.polyu.hk; 202401004@hzxy.edu.cn; shenxy0385@shu.edu.cn; zhangheping@mails.ccnu.edu.cn). (Corresponding author: Peiwen Yang.)

Spatiotemporal calibration is fundamental to robotic vision because transmission time offsets often can lead to temporal desynchronization between image sequences and robot motions [14]. At high robot speeds, time offsets significantly reduce measurement accuracy [15]. In addition, software-based temporal synchronization offers a hardware-agnostic solution for spatiotemporal data association. Additionally, hand-eye extrinsic parameters may vary during manufacturing operations. Calibrating these parameters online will greatly improve efficiency.

*Research background and previous research findings*: Temporal calibration is a critical problem in robotic vision systems, a field that has garnered significant research attention. Time-offset estimation methods are primarily divided into two categories: filter-based and optimization-based methods. Li [16] proposed an online approach aiming at estimating the time-offset between the camera and inertial measurement unit (IMU) [17-19] during EKF-based vision-aided inertial navigation. The EKF-based methods continue to be widely adopted owing to their high efficiency. However, EKF-based methods require careful design of initial states and noise parameters. In contrast, optimization-based methods have been extensively utilized in time synchronization with multiple sensors. Kelly [20] and Wu [21] approached time-offset estimation as an inherent problem, and estimated time-offset based on the iterative closest point (ICP) algorithm [22-24]. In addition, there are many methods based on visual-inertial odometry (VIO) [25-27] to calibrate the time delay or external parameters between the camera and IMU. However, at the machining site, the solution of visual odometry is prone to failure due to the deficiency of the visual features of the weld images. Although many methods have been proposed to acquire higher efficiency. In the field of industrial robots, the calibration of time-offset remains underexplored and addressed.

Additionally, the hand-eye calibration of industrial robots has been extensively explored in [15, 28, 29]. Yin [30] calibrated the LVS within a robotic vision system using a standard sphere. In [31], an efficient calibration method for line-structured light vision sensors in robotic eye-in-hand systems is proposed based on the checkerboard's observations. The method [34] was also based on high-precision checkerboard, which improved calibration efficiency and achieved high accuracy automatically. This method provides calibration accuracy performance comparable to commercial robot factory tools. Some researchers utilized specialized calibration objects to formulate constraint equations via point cloud registration algorithm [32, 33]. For instance, in [35], a calibration method based on reconstruction using arbitrary objects was introduced. Pose-based and point-based methods on compliance assembly robots were compared in [36]. While hand-eye calibration well-developed; few studies have combined it with temporal calibration. Moreover, the calibration process typically demands a tedious manual teaching procedure.

*Motivation*: We now briefly describe the motivation of this paper here. All the above methods require collecting a large amount of data and fail to integrate temporal calibration. What's more, these methods require stopping the machine and tedious teaching processes. The previous discussion indicates that the spatiotemporal calibration of the LVS in the hand-eye system has not yet been effectively solved. Therefore, to ensure accuracy and efficiency, reliable spatiotemporal calibration technology for LVS-based hand-eye systems is urgent. In this study, we harness the externally guided motion (EGM) interface [37], which enables high frequency and low delay control of ABB industrial robots.

*Contribution*: In this article, a spatiotemporal calibration method for LVS in the hand-eye system based on straight-line constraints is proposed. Without teaching the robot, the robot scans a straight-line fillet weld placed nearby and scanned at different postures along an S-shaped path. Specifically, the data process can be performed during the manufacturing process, reducing the risks of manufacturing failures. The straight-line constraint is used to build a nonlinear optimization model. The main contributions are as follows:

1) Considering the time-offset and hand-eye parameters, an LVS measurement model is improved.
2) Using straight-line constraints, a spatiotemporal calibration method for a hand-eye robotic system is proposed. We model the straight-line fillet weld using Plucker coordinates [38, 39]. Since the weld feature point measurements are located on a spatial straight-line, a nonlinear optimization model is constructed based on the straight-line constraints. In addition, the Jacobians of the error function with respect to the parameters to be estimated are given. Crucially, we represent both straight-line weldment parameters and hand-eye extrinsic relationships using Lie Group theory. By formulating the optimization problem on this manifold, we implement Levenberg-Marquardt algorithm (LMA) [40] that inherently reduces linearization errors.
3) Finally, experiments are performed to verify the feasibility and accuracy of the proposed method. The calibration results demonstrate comparable or superior accuracy to existing state-of-the-art methods when validated against ground truth. The root mean square error (RMSE) and mean absolute error (MAE) of the weld position scanning are less than 0.2 mm. To benefit the research community, we open-source all the datasets and code.

## II. Methodology

The operator $(\cdot)^{\wedge}$ or $\lfloor \cdot \times \rfloor$ represents the operation that transforms a vector into a skew-symmetric matrix. ${}^{A}_{B}\mathbf{T}$ in special Euclidean group $SE(3)$ is defined as the homogeneous transformation matrix, representing poses of the coordinate system $\{B\}$ in the coordinate system $\{A\}$, with the form of

$$ {}^{A}_{B}\mathbf{T} = \begin{bmatrix} {}^{A}_{B}\mathbf{R} & {}^{A}\mathbf{P}_{B} \\ \mathbf{0} & 1 \end{bmatrix} \tag{1} $$

where ${}^{A}_{B}\mathbf{R} \in SO(3)$ and ${}^{A}\mathbf{P}_{B} \in \mathbf{R}^{3}$ are rotation and translation from the frame $\{F_B\}$ to the frame $\{F_A\}$.

*This section proposes an improved LVS model, integrating time-offset. Then, the cost problem of the proposed calibration method and Jacobian matrices are given.* Fig. 1 introduces the LVS model considering time-offset in the hand-eye system and presents the spatiotemporal calibration scheme. The LVS-based measurement system contains a base coordinate system (BCS) [2], a camera coordinate system (CCS) {C}, and an end-effector coordinate system (ECS) $\{E\}$.

As shown in Fig. 1, a fillet weld is placed on a workbench during the calibration process, waiting to be scanned by a 6DoF industrial robot. There are hand-eye parameters between the CCS and the ESC. In addition, due to the transmission delay, there is a time-offset between the images from the LVS and the motion data from the industrial robot. The robot follows the S-shape path to scan a fillet weld, and visual features can be converted using coordinate transformation. The $i$ -th weld feature $\mathbf{f}_i^t$ in the images at the timestamp $t$ can be converted into its position ${}^{C}\mathbf{P}_{f_i}^{t}$ in the CCS using the LVS measurement model. The pose of the ECS in the BCS can be obtained using the EGM of the industrial robot. $t_d$ is the time-offset and ${}^{E}\mathbf{T}_{C}$ represents the hand-eye transformation. ${}^{B}\mathbf{T}_{E}^{t-t_d}$ represents the pose of the end-effector at the timestamp $t - t_d$, which corresponds to the image captured by the camera at time $t$. However, because the camera data is lagged, the true correspondence is the pose at timestamp $t - t_d$. Therefore, we propose a unifying method to complete the spatial and temporal calibration unified. In addition, teaching free method is vital to eliminate errors caused by production and can assure equipment measurement accuracy.

### A. Time-offset and LVS measurement model

The receiving timestamp of the image is not equal to the sampling time; this will result in time-offset between the image and robot motion. Compared to the industrial camera, the EGM module time-offset can be ignored. The EGM time $t_{EGM}$ and image time $t_{CAM}$ are synchronized by software. $t_{CR}$ and $t_{ER}$ represent the camera receiving and robot motion receiving times, respectively. Therefore, as illustrated in Fig. 2, based on the system clock of the computer, the trigger time of the image $i$ equals $t_{CR}^{i}-t_d$, where $t_{CR}$ is the image receiving time. When the condition $t_{ER}^{j} \le t_{CR}^{i}-t_d \le t_{ER}^{j+1}$ is satisfied, the robot poses corresponding to the image $i$ can be obtained through spherical linear interpolation from $\mathbf{P}^{j}$ and $\mathbf{P}^{j+1}$. Due to hardware limitations, the frequency of this camera is 36Hz. However, due to limitations in data volume and data interfaces, time-offset is inevitable and frequency independent.

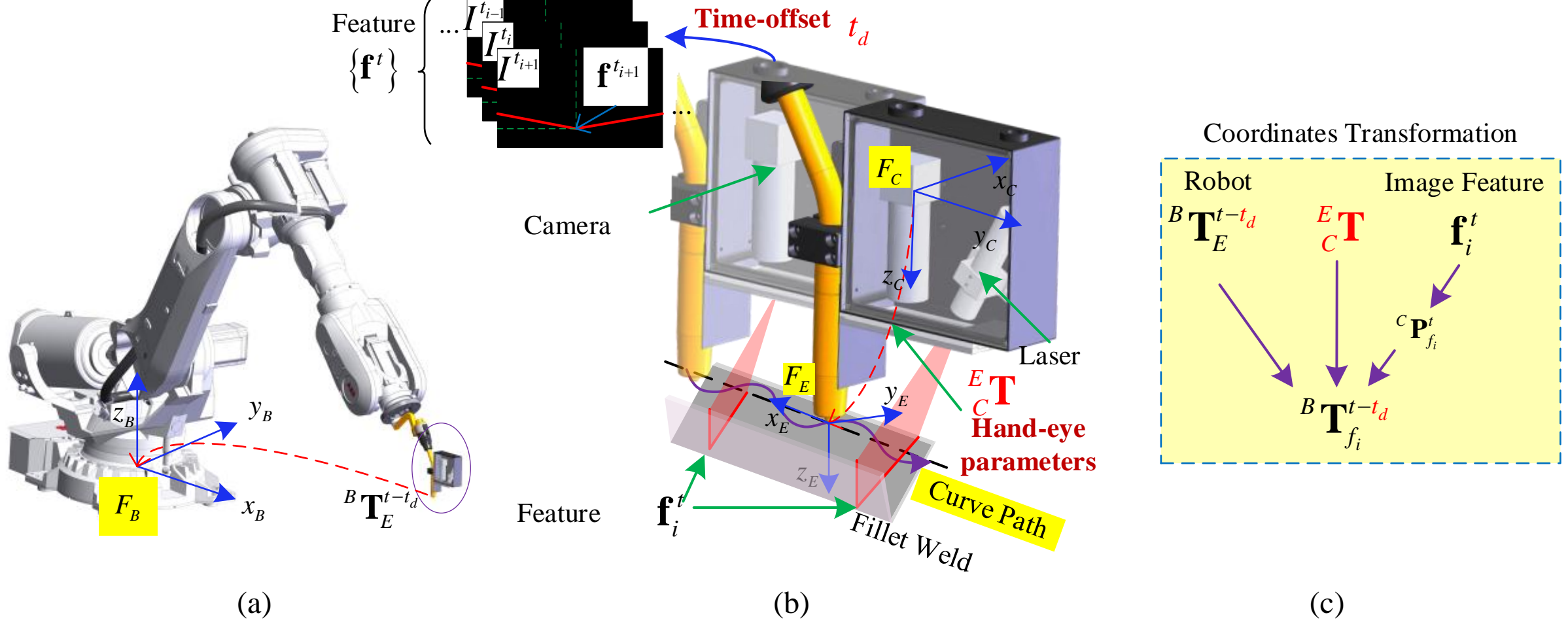

Fig. 1. An illustration of LVS spatiotemporal calibration in the hand-eye system. (a) The industrial hand-eye manufacturing system with an LVS sensor. (b) The calibration scheme and the unknown parameters, time-offset and hand-eye parameters. (c) The coordinates transformation between the robot poses and weldment images' feature.

When an LVS is scanning a fillet weld, the intersection points of two laser stripes are obtained by the image processing algorithm as feature points, which are used to calculate the spatial coordinates of discrete points on a fillet weld. Given the facts above, an error model based on straight-line constraint is constructed to optimize the Plucker coordinates of the straight-line, hand-eye parameters, and the time-offset.

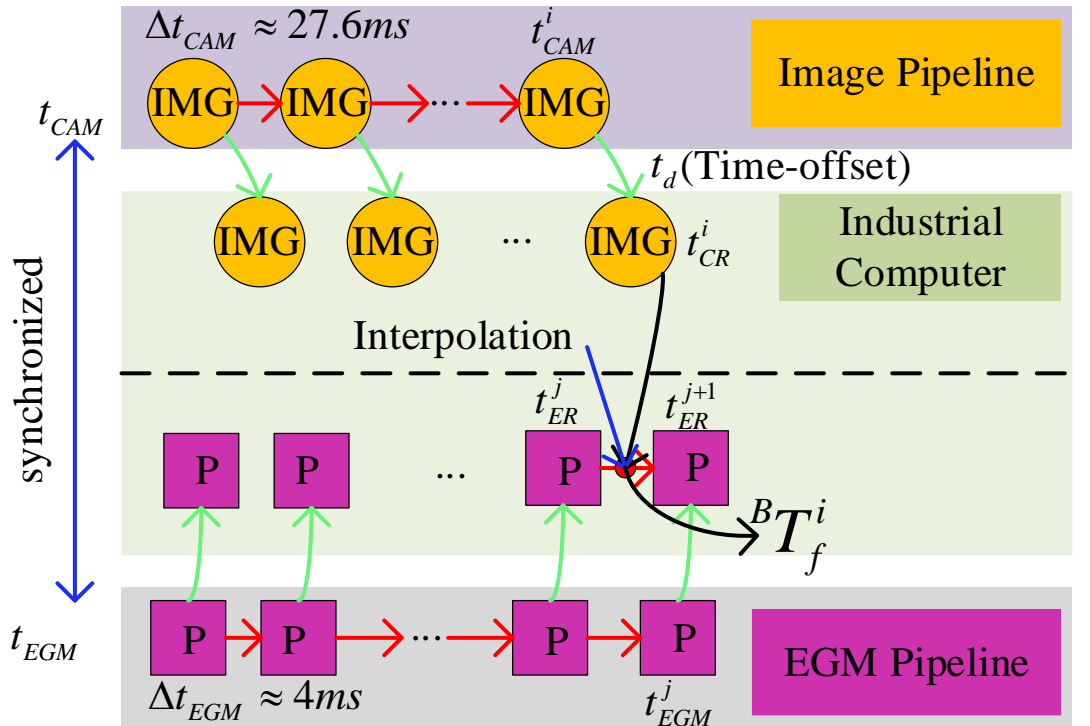

Fig. 2. The EGM (robot motion) pipeline (purple square) and image pipeline (orange circle). Every image has a time-offset $t_d$.

Without time-offset, the transformation relation between the laser stripe image features $\mathbf{f}^t=\left[u^t, v^t, 1\right]^T$ (in homogeneous form) and its 3D coordinates ${}^{B}\mathbf{P}_f^t=\left[{}^{B}\mathbf{P}_x^t, {}^{B}\mathbf{P}_y^t, {}^{B}\mathbf{P}_z^t, 1\right]^T$ (in homogeneous form) at the time $t$ is shown as follows:

$$ {}^{B}\mathbf{P}_f^t = {}^{B}_{E}\mathbf{T}^t \cdot {}^{E}_{C}\mathbf{T} \cdot \mathrm{G}(\mathbf{f}^t) \tag{2} $$

where $G(\mathbf{f}^t)$ denotes the transformation that maps a laser stripe point in the image pixel coordinate system (IPCS) to a 3D coordinate ${}^{C}\mathbf{P}_f^t=\left[{}^{C}P_x^t, {}^{C}P_y^t, {}^{C}P_z^t, 1\right]^T$ in the CCS. The ${}^{E}_{C}\mathbf{T}$ denotes hand-eye parameters, and $G(\mathbf{f}^t)$ is the mapping function.

When the time-offset is considered, the image at the timestamp $t$ corresponds to the robot end-effector's pose ${}^{B}_{E}\mathbf{T}^{t-t_d}$ at the timestamp $t-t_d$. Therefore, the measurement model of the LVS in the hand-eye system considering the time-offset is as follows:

$$ {}^{B}\mathbf{P}_f^{t-t_d} = {}^{B}_{E}\mathbf{R}^{t-t_d}\left({}^{E}_{C}\mathbf{R}\left({}^{C}\mathbf{P}_f^t\right)_{(1:3)} + {}^{E}\mathbf{P}_C\right) + {}^{B}\mathbf{P}_E^{t-t_d} + \mathbf{w}_k \tag{3} $$

where $\{\mathbf{w}_k\}$ is the zero-mean white Gaussian noise process. The robot end-effector's rotation and position are ${}^{B}_{E}\mathbf{R}^{t-t_d}$ and ${}^{B}\mathbf{P}_E^{t-t_d}$, respectively, at the timestamp $t-t_d$, and hand-eye parameters include ${}^{E}_{C}\mathbf{R}$ and ${}^{E}\mathbf{P}_C$.

### *B. Kinematics*

In this work, the robot is assumed to move at constant angular and linear velocities over a brief interval. Therefore, the robot's pose at an arbitrary time instant can be interpolated using its nearest robot poses. The end-effector's angular velocity ${}^{E}\boldsymbol{\omega}^{t_i}$ and linear velocity ${}^{B}\mathbf{v}_E^{t_i}$ at $t_i \in \left[t_j, t_{j+1}\right)$ are approximately,

$$ \begin{aligned} {}^{E}\boldsymbol{\omega}^{t_i} &= \mathrm{Log}({}^{B}_{E}\mathbf{R}_j^T \cdot {}^{B}_{E}\mathbf{R}_{j+1})/(t_{j+1}-t_j) \\ {}^{B}\mathbf{v}_E^{t_i} &= ({}^{B}\mathbf{P}_E^{j+1} - {}^{B}\mathbf{P}_E^{j})/(t_{j+1}-t_j) \end{aligned} \tag{4} $$

where the robot's position at the timestamp $t_j$ and $t_{j+1}$ are ${}^{B}\mathbf{P}_E^{j}$ and ${}^{B}\mathbf{P}_E^{j+1}$, and the robot's rotation at the timestamp $t_j$ and $t_{j+1}$ are ${}^{B}_{E}\mathbf{R}_j$ and ${}^{B}_{E}\mathbf{R}_{j+1}$. $t_j$ and $t_{j+1}$ is the nearest timestamp about $t_i$.

The interpolation of robot rotation and position at $t-t_d$ ($t_j \le t-t_d < t_{j+1}$) are denoted as follows [18],

$$\begin{aligned} {}^{B}\mathbf{R}_{E}^{t-t_d} &= {}^{B}\mathbf{R}_{E}^{t_j} \cdot \mathrm{Exp}\left( {}^{E}\boldsymbol{\omega}^{t_i} (t-t_d-t_j) \right) \\ {}^{B}\mathbf{P}_{E}^{t-t_d} &= {}^{B}\mathbf{P}_{E}^{t_j} + {}^{B}\mathbf{v}_{E}^{t_i} \cdot \left( t-t_d-t_j \right) \end{aligned} \tag{5}$$

The Jacobian of the interpolated rotation and position concerning $t_d$ are as follows:

$$\begin{aligned} \frac{\partial\, {}^{B}\mathbf{R}_{E}^{t-t_d}}{\partial t_d} &= -{}^{B}\mathbf{R}_{E}^{t_j} \cdot {}^{E}\boldsymbol{\omega}^{t_i} \\ \frac{\partial\, {}^{B}\mathbf{P}_{E}^{t-t_d}}{\partial t_d} &= -{}^{B}\mathbf{v}_{E}^{t_i} \end{aligned} \tag{6}$$

Therefore, according to Eq. (3), Eq. (6), and chain rule for differentiation, the Jacobian of the weld feature points' position concerning $t_d$ is:

$$\begin{aligned} \frac{\partial\, {}^{B}\mathbf{P}_{f_i}^{t-t_d}}{\partial t_d} &= \frac{\partial\, {}^{B}\mathbf{P}_{f_i}^{t-t_d}}{\partial\, {}^{B}\mathbf{R}_{E}^{t-t_d}} \cdot \frac{\partial\, {}^{B}\mathbf{R}_{E}^{t-t_d}}{\partial t_d} + \frac{\partial\, {}^{B}\mathbf{P}_{f_i}^{t-t_d}}{\partial\, {}^{B}\mathbf{P}_{E}^{t-t_d}} \cdot \frac{\partial\, {}^{B}\mathbf{P}_{E}^{t-t_d}}{\partial t_d} \\ &= -\left( {}^{B}\mathbf{R}_{E}^{t-t_d}\, {}^{E}\mathbf{P}_{f_i} \right)^{\wedge}\, {}^{B}\mathbf{R}_{E}^{t_i} \cdot {}^{E}\boldsymbol{\omega}^{t_i} - {}^{B}\mathbf{v}_{E}^{t_i} \end{aligned} \tag{7}$$

C. *Line constraint and parameters optimization*

The straight-line's Plucker coordinates $[\mathbf{n}, \mathbf{v}]$ contain a unit vector $\mathbf{v}$ and a moment $\mathbf{n}$ with $\mathbf{n} = \mathbf{p}_0 \times \mathbf{v}$ for a given point $\mathbf{p}_0$ on the line. The minimum representation $(\mathbf{U} \quad m) \in SO(3) \times \mathbf{R}$ can be obtained from the Plucker coordinates [38, 39]:

$$\mathbf{L}(\phi) = [\mathbf{v} \mid \mathbf{n}] = \begin{bmatrix} \mathbf{v} & \dfrac{\mathbf{n}}{|\mathbf{n}|} & \dfrac{\mathbf{v}\times\mathbf{n}}{|\mathbf{n}|} \end{bmatrix} \begin{bmatrix} \mathbf{1} & \mathbf{0} \\ \mathbf{0} & |\mathbf{n}| \\ \mathbf{0} & \mathbf{0} \end{bmatrix} \sim \mathbf{U}(\boldsymbol{\omega})\mathbf{W}(m) \tag{8}$$

$$\mathbf{U}(\boldsymbol{\omega}) = \begin{bmatrix} \mathbf{v} & \dfrac{\mathbf{n}}{|\mathbf{n}|} & \dfrac{\mathbf{v}\times\mathbf{n}}{|\mathbf{n}|} \end{bmatrix} = \begin{bmatrix} \mathbf{u_1} & \mathbf{u_2} & \mathbf{u_3} \end{bmatrix} \in SO3, m = |\mathbf{n}|$$

Therefore, a 3D straight-line $\mathbf{L} \in \mathbf{R}^6$ has a minimum parameterization $\phi = [\boldsymbol{\omega} \quad m]^T \in \mathbf{R}^4 (\boldsymbol{\omega} \in so3, m \in \mathbf{R})$. The Plucker coordinates can be converted from the orthonormal representation by,

$$\begin{cases} \mathbf{n} = \mathbf{u_2} \cdot |\mathbf{n}| = \mathbf{U}(\boldsymbol{\omega})\mathbf{e}_2 \cdot m = m\,\mathrm{Exp}(\boldsymbol{\omega})\mathbf{e}_2 \\ \mathbf{v} = \mathbf{u_1} = \mathbf{U}(\boldsymbol{\omega})\mathbf{e}_1 = \mathrm{Exp}(\boldsymbol{\omega})\mathbf{e}_1 \end{cases} \tag{9}$$

where $\mathbf{e}_1 = [1,0,0]^{\mathrm{T}}$ and $\mathbf{e}_2 = [0,1,0]^{\mathrm{T}}$.

The cross product of the vector $\left( {}^{B}\mathbf{P}_{f}^{t-t_d} - \mathbf{P}_0 \right)$ and the vector $\mathbf{v}$ can be constructed as the distance vector (error vector). Therefore, the distance (cost function) between a 3D point ${}^{B}\mathbf{P}_{f_i}^{t-t_d}$ to a line $\mathbf{L}$ is as follows:

$$\begin{aligned} & \mathrm{F}^t\left( {}^{B}\mathbf{P}_{f}^{t-t_d}, \mathbf{L} \right) \\ &= \left( {}^{B}\mathbf{P}_{f}^{t-t_d} - \mathbf{P}_0 \right) \times \mathbf{v} = {}^{B}\mathbf{P}_{f}^{t-t_d} \times \mathbf{v} - \mathbf{P}_0 \times \mathbf{v} \\ &= {}^{B}\mathbf{P}_{f}^{t-t_d} \times \mathbf{v} - \mathbf{n} \\ &= \left( {}^{B}_{E}\mathbf{R}^{t-t_d} \left( {}^{E}_{C}\mathbf{R}\, {}^{C}\mathbf{P}_{f}^{t} + {}^{E}\mathbf{P}_{C} \right) + {}^{B}\mathbf{P}_{E}^{t-t_d} \right) \times \mathbf{v} - \mathbf{n} \end{aligned} \tag{10}$$

According to Eq. (8) and Eq. (10), given a small perturbation $\delta\phi = [\delta\boldsymbol{\omega} \quad \delta m]$ for the straight-line parameters, the Jacobian matrix of $\mathrm{F}^t$ concerning $\phi$ is as follows:

$$\begin{aligned} \mathbf{J}_{\phi}^{t} &= \frac{\partial \mathrm{F}^t}{\partial \phi} = \frac{\partial\, {}^{B}\mathbf{P}_{f}^{t-t_d} \times \mathbf{v}}{\partial \phi} + \frac{\partial -\mathbf{n}}{\partial \phi} = -{}^{B}\mathbf{P}_{f}^{t-t_d} \times \frac{\partial \mathbf{v}}{\partial \phi} - \frac{\partial \mathbf{n}}{\partial \phi} \\ &= -{}^{B}\mathbf{P}_{f}^{t-t_d} \left[ \left( \mathbf{U}(\boldsymbol{\omega})\mathbf{e}_1 \right)^{\wedge} \quad \mathbf{0}_{3\times1} \right] - \left[ m\left( \mathbf{U}(\boldsymbol{\omega})\mathbf{e}_2 \right)^{\wedge} \quad \mathbf{U}(\boldsymbol{\omega})\mathbf{e}_2 \right] \\ &= -\left[ {}^{B}\mathbf{P}_{f}^{t-t_d} \times \mathbf{u}_1^{\wedge} + m\mathbf{u}_2^{\wedge} \quad \mid \quad \mathbf{u}_2 \right] \end{aligned} \tag{11}$$

where $\mathbf{u}_1 = \mathrm{Exp}(\boldsymbol{\omega}) \cdot \mathbf{e}_1$ and $\mathbf{u}_2 = \mathrm{Exp}(\boldsymbol{\omega}) \cdot \mathbf{e}_2$.

Furthermore, according to Eq. (7) and Eq. (10), the Jacobian matrix of $\mathrm{F}^t$ concerning time-offset $t_d$ is as follows:

$$\begin{aligned} \mathbf{J}_{t_d}^{t} &= \frac{\partial \mathrm{F}^t}{\partial t_d} = \frac{\partial \mathrm{F}^t}{\partial\, {}^{B}\mathbf{P}_{f}^{t-t_d}} \cdot \frac{\partial\, {}^{B}\mathbf{P}_{f}^{t-t_d}}{\partial t_d} \\ &= \lfloor -\mathbf{v}\times \rfloor \cdot \left\{ -\left( {}^{B}_{E}\mathbf{R}^{t-t_d}\, {}^{E}\mathbf{P}_{f} \right)^{\wedge}\, {}^{B}_{E}\mathbf{R}^{t_i} \cdot \boldsymbol{\omega}_{E}^{t_i} + {}^{B}\mathbf{v}_{E}^{t_i} \right\} \end{aligned} \tag{12}$$

If the robot remains in a constant attitude while scanning the fillet weld, the rotation part $-\left( {}^{B}\mathbf{R}_{E}^{t-t_d}\, {}^{E}\mathbf{P}_{f_i} \right)^{\wedge}\, {}^{B}\mathbf{R}_{E}^{t_i} \cdot {}^{E}\boldsymbol{\omega}^{t_i}$ of the Jacobian matrix $\partial \mathrm{F}^t / \partial t_d$ will be zero and the Jacobian matrix $\partial \mathrm{F}^t / \partial t_d$ becomes:

$$\mathbf{J}_{t_d}^{t} = \lfloor -\mathbf{v}\times \rfloor \cdot {}^{B}\mathbf{v}_{E}^{t_i} \tag{13}$$

Time offset calibration requires dynamic excitation to break unobservability (such as changing motion speed). The robot end-effector's velocity is required to be time-varying to ensure $\Delta t$'s observability. Therefore, the S-shaped or C-shaped path is suitable for calibration. As shown in Fig. 2 (b), the robot end-effector's path is represented by a purple curved line, subsequently the LVS is driven to measure a fillet weld according to the pre-set path. Then, the feature coordinates $\mathbf{f}^t$ of the laser stripe in the CCS are transformed into the BCS.

*This method can not only calibrate the time-offset of the camera but also calibrate the hand-eye parameters.* The Jacobian matrix of $\mathrm{F}^t$ concerning ${}^{E}_{C}\mathbf{R}$ is as follows:

$$\begin{aligned} \mathbf{J}_{R}^{t} &= \frac{\partial \mathrm{F}^t}{\partial\, {}^{E}_{C}\mathbf{R}} = \frac{\partial \mathrm{F}^t}{\partial\, {}^{B}\mathbf{P}_{f}^{t-t_d}} \cdot \frac{\partial\, {}^{B}\mathbf{P}_{f}^{t-t_d}}{\partial\, {}^{E}_{C}\mathbf{R}} \\ &= -\lfloor -\mathbf{v}\times \rfloor \cdot {}^{B}_{E}\mathbf{R}^{t-t_d} \cdot \left( {}^{E}_{C}\mathbf{R} \cdot {}^{C}\mathbf{P}_{f}^{t} \right)^{\wedge} \end{aligned} \tag{14}$$

The Jacobian matrix of $\mathrm{F}^t$ concerning ${}^{E}\mathbf{P}_{C}$ is as follows:

$$\mathbf{J}_{P}^{t} = \frac{\partial \mathrm{F}^t}{\partial\, {}^{E}\mathbf{P}_{C}} = \frac{\partial \mathrm{F}^t}{\partial\, {}^{B}\mathbf{P}_{f}^{t-t_d}} \cdot \frac{\partial\, {}^{B}\mathbf{P}_{f}^{t-t_d}}{\partial\, {}^{E}\mathbf{P}_{C}} = \lfloor -\mathbf{v}\times \rfloor \cdot {}^{B}_{E}\mathbf{R}^{t-t_d} \tag{15}$$

The state is $\mathbf{x} = \{ t_d, {}^{B}\phi, {}^{E}\mathbf{P}_{C}, {}^{E}_{C}\mathbf{P} \}$. $t_d$ is the time-offset, and ${}^{B}\phi$ is the minimal representation of the line. ${}^{E}\mathbf{P}_{C}$ and ${}^{E}_{C}\mathbf{R}$ are the translation and rotation parameters in the hand-eye system. The objective function is constructed using the sum of squares of errors as follows:

$$\hat{\mathbf{x}} = \arg\min \sum_{t}^{S} \rho\left(\left\|F^{t}(\mathbf{x})\right\|_{2}^{2}\right) \tag{16}$$

where $S$ is the timestamp sequences of all images, and $\rho(s)$ is the Huber norm [40]. We perform LMA to optimize the straight-line, hand-eye parameters, and time-offset. Although the LMA may be sensitive to initial values, the least squares (LS) method can be used to obtain an approximate hand eye parameter. And the camera usually has a factory value that can be used as an initial value.

The increment of state $\delta\mathbf{x} = \{\delta t_d, \delta^{B}\phi, \delta^{E}\mathbf{P}_C, \delta^{E}_{C}\mathbf{R}\}$ can be computed using,

$$\begin{gathered} \delta\mathbf{x} = -(\mathbf{J}(\mathbf{x})^{\mathrm{T}}\mathbf{J}(x) + \mu\mathbf{I})^{-1}\mathbf{J}(\mathbf{x})^{\mathrm{T}}F(\mathbf{x}), \\ \mathbf{J} = \left[\mathbf{J}^{t}_{t_d}, \mathbf{J}^{t}_{\phi}, \mathbf{J}^{t}_{P}, \mathbf{J}^{t}_{R}\right] \end{gathered} \tag{17}$$

In each iteration, the unknown parameters $\mathbf{x}$ are updated by

$$\begin{cases} \mathbf{U}' = \mathrm{Exp}(\delta\boldsymbol{\omega}) \cdot \mathrm{U}(\boldsymbol{\omega}) \\ m' = m + \delta m \\ t'_d = t_d + \delta t_d \\ {}^{E}\mathbf{P}'_C = {}^{E}\mathbf{P}_C + \delta^{E}\mathbf{P}_C \\ {}^{E}_{C}\mathbf{R}' = \mathrm{Exp}(\delta^{E}_{C}\phi) \cdot {}^{E}_{C}\mathbf{R} \end{cases} \tag{18}$$

In the end, the optimization process will be stopped when $\delta\mathbf{x}$ is less than a given value $\varepsilon$ or the number of iterations is greater than $N$.

## III. Instruments and Experimental Setup

As shown in Fig. 3 (a), the experiment configuration used in this study comprises an industrial personal computer (IPC), a self-made LVS, and an industrial robot (ABB, IRB6700). The LVS is rigidly mounted on a welding torch and thus forms a hand-eye system. The LVS transmits the images to the IPC at 33 Hz. In addition, the IPC controls the robotic path using the EGM interface, which can retrieve robot feedback every 4 milliseconds. The industrial camera in the LVS is a CMOS camera (DAHENG IMAGING, MER2-503-23GM/C-P, China) with a resolution of 2448 × 2000 pixels. To provide more details and promote the development of the industry, we have open-sourced relevant hardware solutions, datasets, and codes.

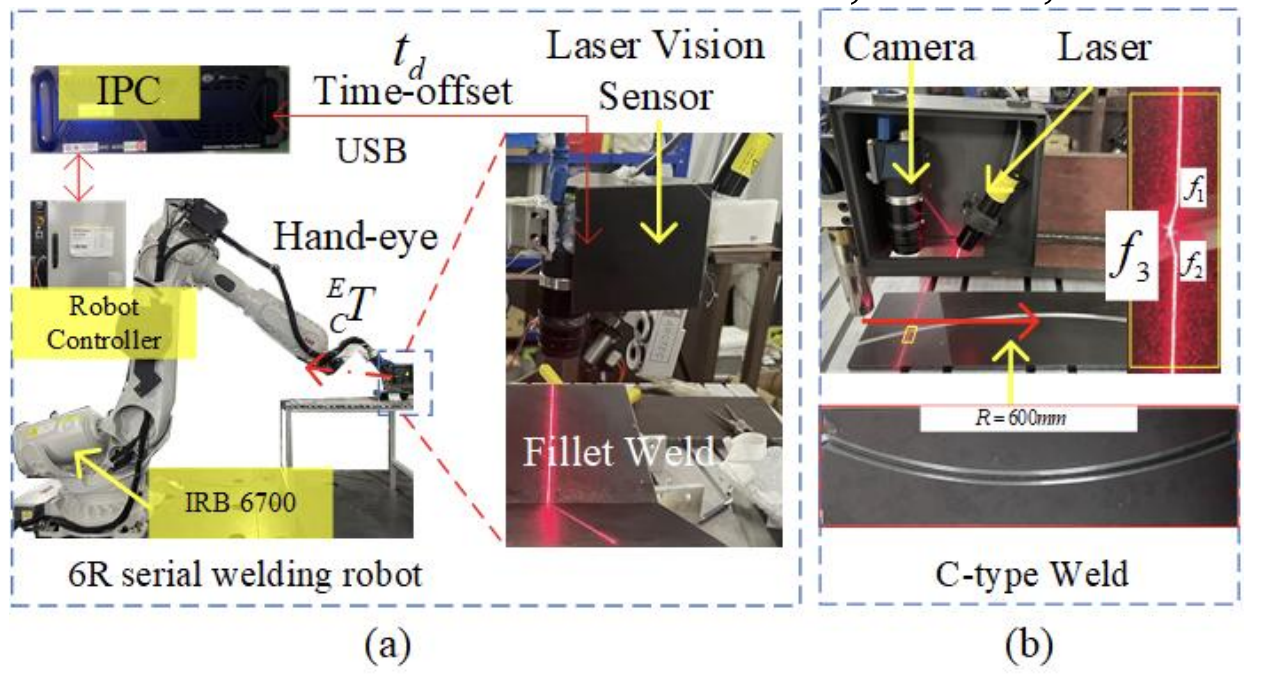

Fig. 3. Experimental equipment configuration. (a) The industrial robot and the self-made LVS. (b) The C-type weld scanning for accuracy validation.

## IV. Experimental Context, Results, and Discussion

### A. Real-world calibration experiments

The proposed method is evaluated using a line-straight weld. The calibrated parameters are compared with the ground truth via offline calibration: time-offset is obtained from the method in [27] and hand-eye parameters are obtained from the robot internal calibration method. All these methods have demonstrated calibration performance and can be used as the ground truth. (1) *Weld feature detection*: During the calibration process, the weld feature point is detected using machine vision technologies [41-44]. One feature detection example is illustrated in Fig. 4. (2) *Calibration process*: As shown in Fig. 5, the calibration process comprises two steps: data acquisition and parameter optimization. The former completes fillet weld scanning under various attitudes, and the latter constructs an optimization problem based on the calibration data collected in the first stage. Before calculating the objective function, ensure that the robot's pose is interpolated to match the updated image time.

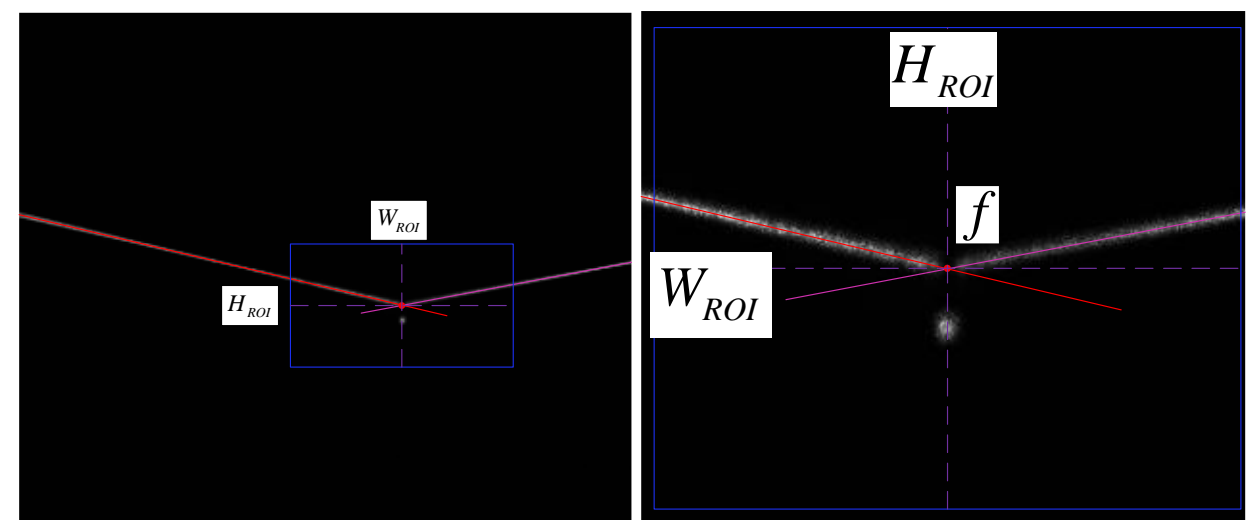

Fig. 4. Fillet weld feature detection.

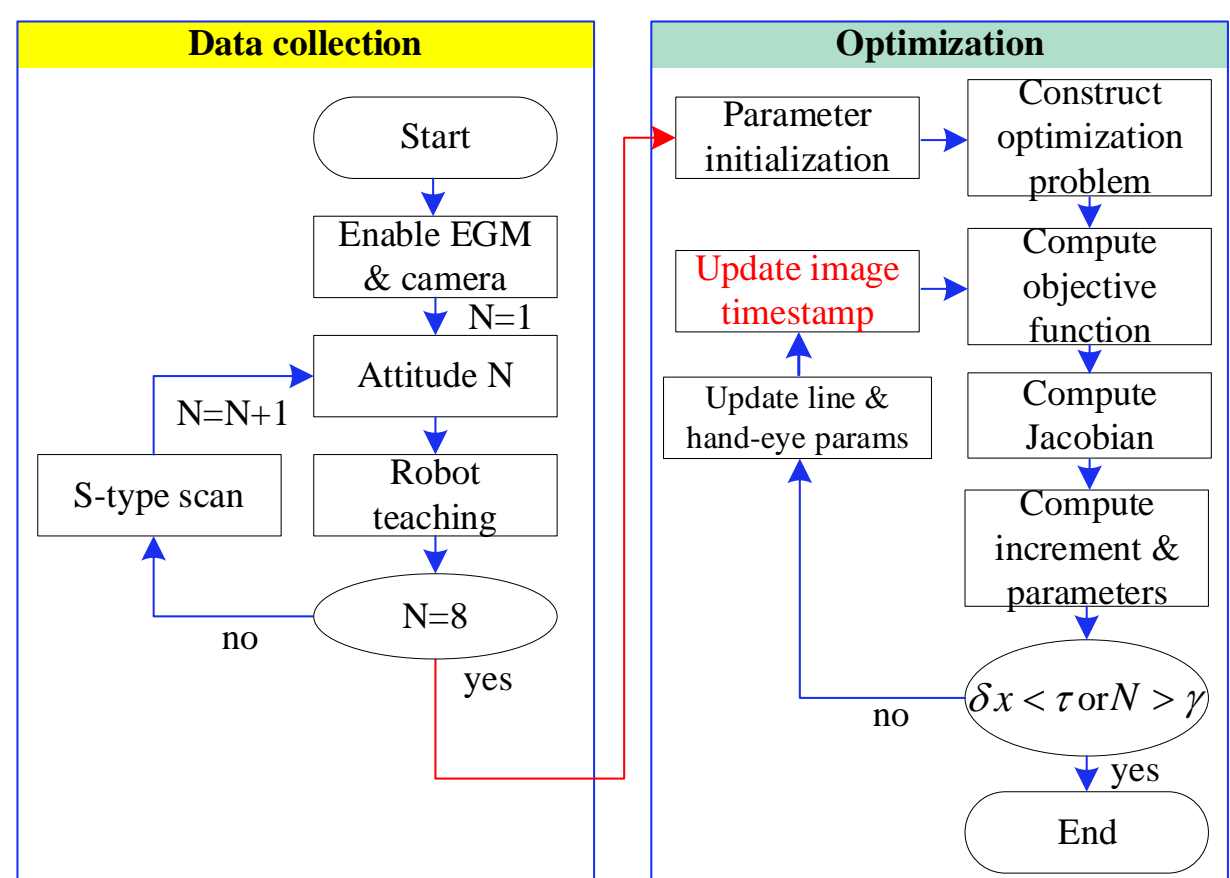

Fig. 5. Flow chart of the calibration process of the proposed method and calibration video can be viewed at the open-sourced page.

The optimization algorithm is implemented by GTSAM. The initial values: $t_d = 0ms$, $\phi = [0.169084 \quad -0.000782 \quad 0.0197593 \quad 1152.86]$. As shown in Fig. 6, the time-offset and RMSE converge effectively. The RMSE reaches the minimum value of 0.32715 mm, and the time-offset converges to 0.01773 seconds. The red curve represents uncalibrated measurements, which exhibit significant deviations due to time offset. Furthermore, the straight-line parameters optimization process is shown in Fig. 7. In addition, Fig. 8 shows the hand-eye parameters' optimization curves, and the hand-eye parameters can also quickly converge to a stable value. All the parameters are evaluated with the ground truth. The time-offset calibration error is less than 1 millisecond. In addition, the error in the translation parameter is within 0.20 mm.

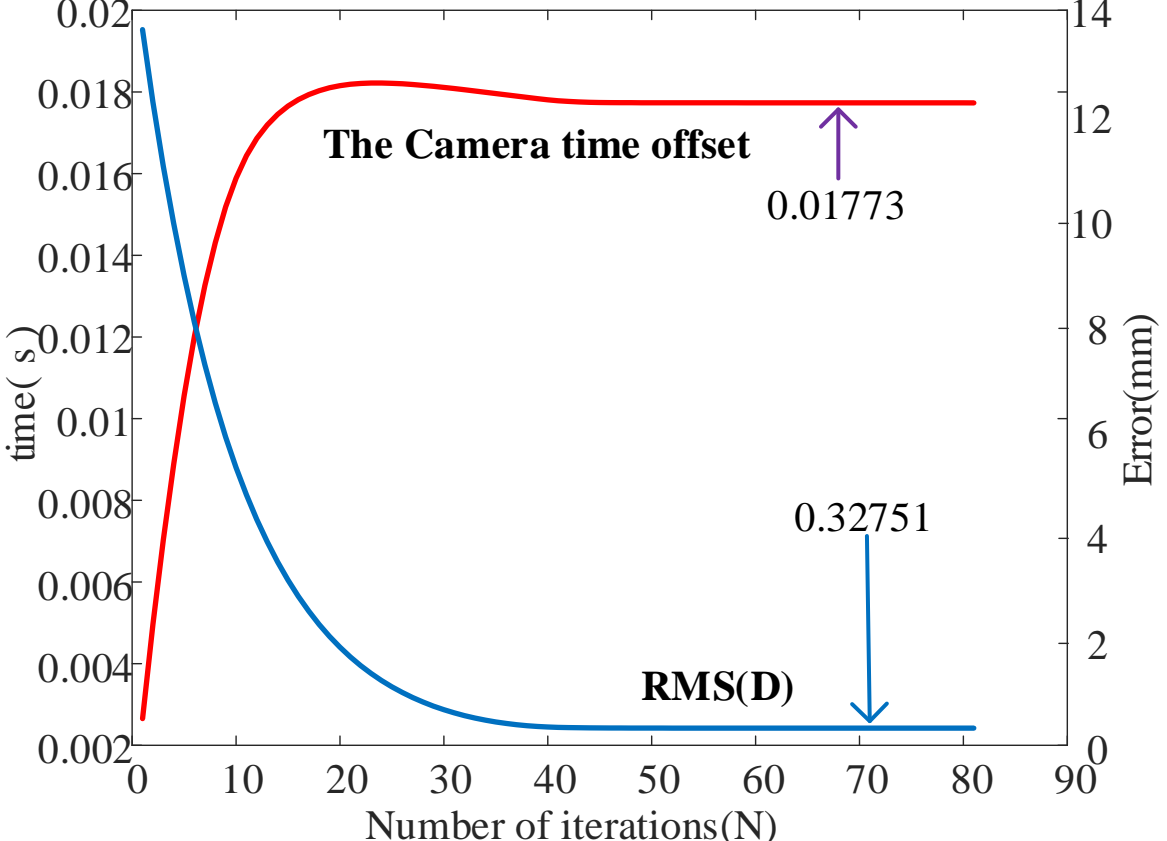

Fig. 6. (a) Values of the time-offset during optimization. (b) Values of the RMS of the point-to-line distance during optimization.

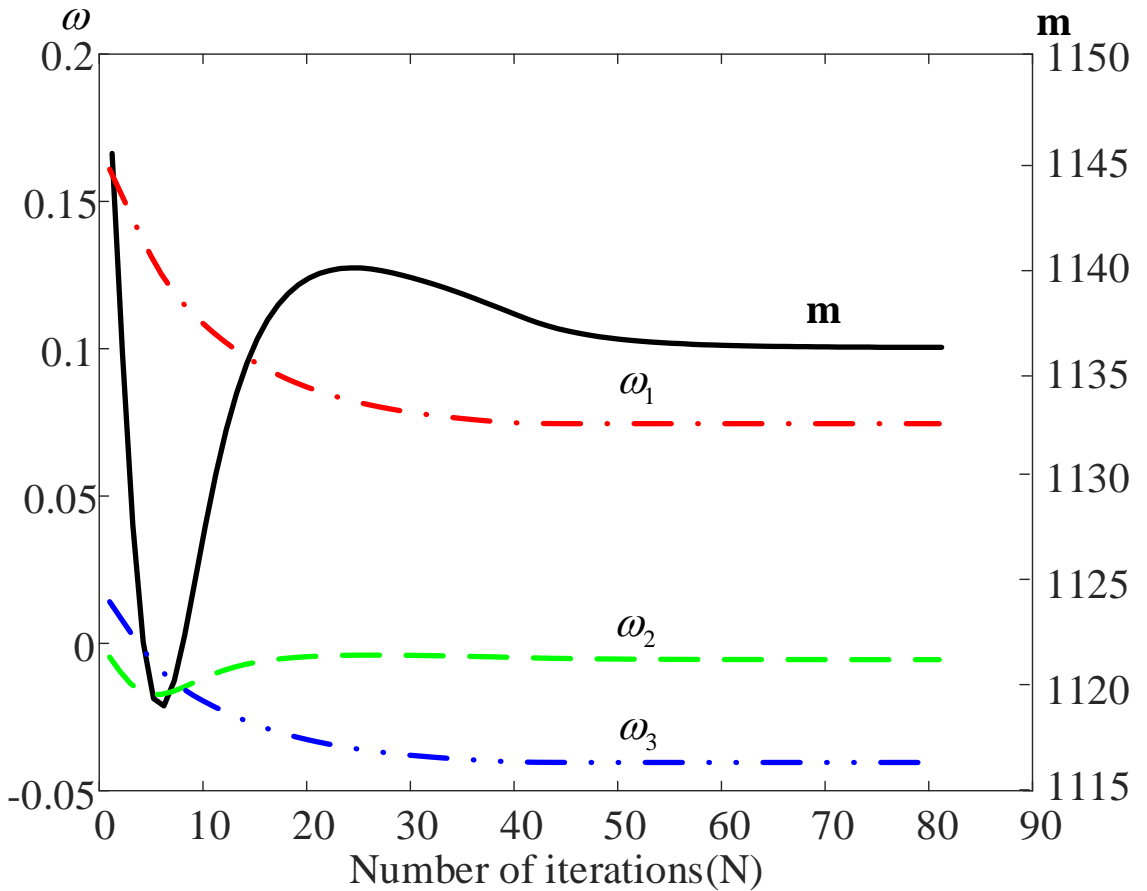

Fig. 7. Values of the line parameters during optimization.

*Time-offset effect*: The weld position's ground truth is obtained from high-precision teaching functions. As shown in Fig. 9, without considering time-offset, the weld's measurement accuracy is increased significantly. Compensating for the time-offset, the RMSE of the weld measurement error is 0.17 mm, which meets the welding requirements.

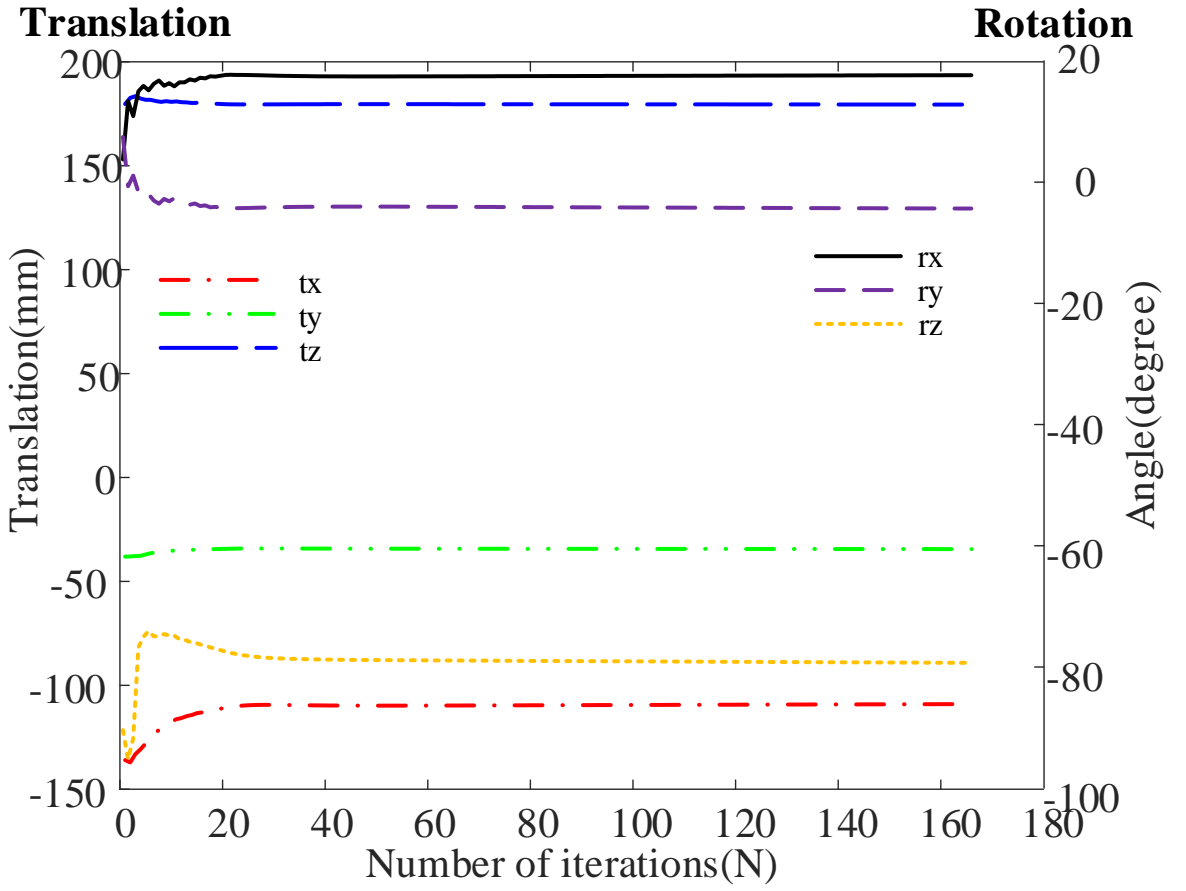

Fig. 8. Values of the hand-eye parameters during optimization.

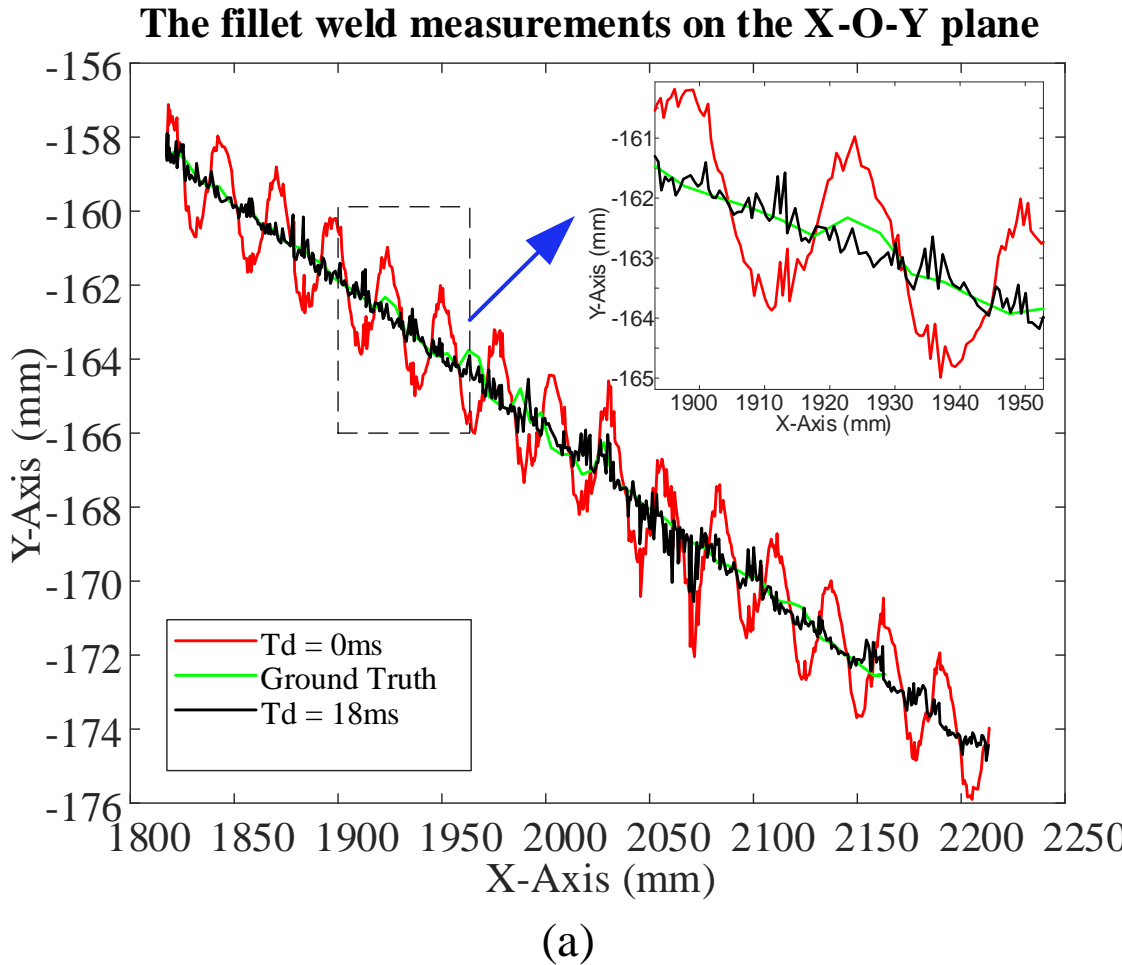

(a)

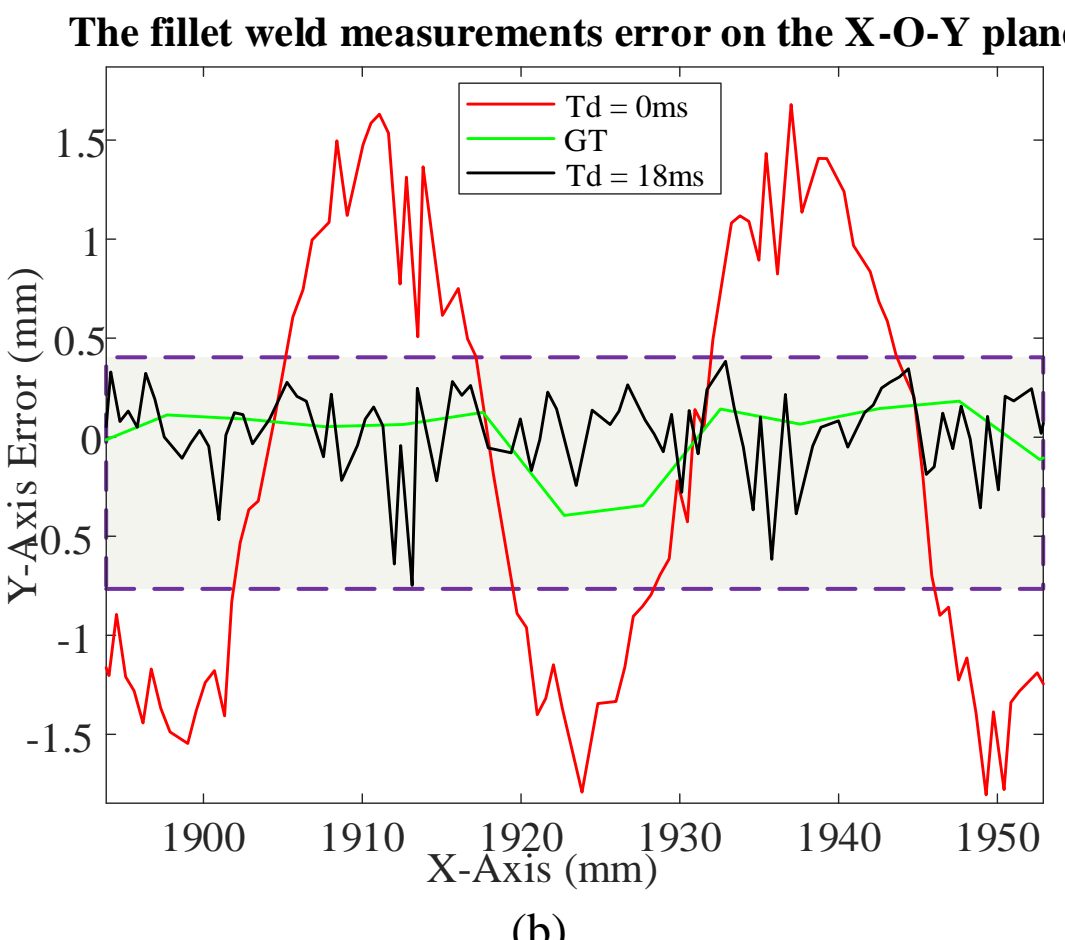

(b)

Fig. 9. The fillet weld measurement without a time-offset calibration is depicted in a red solid line. The fillet weld measurement with time-offset correction is depicted in a black solid line (time-offset calibrated result is $t_d = 18ms$ ). The ground truth of the fillet weld is obtained by static acquisition, depicted in a green solid line.

*Comparisons with existing calibration method*: (1) Time-offset: Compared with the existing temporal calibration method in [27] solving VIO cases very well, however, this method cannot be adapted to LVS. This is because method [24] heavily relies on visual features of the scene, but many factories cannot guarantee sufficient features, even lighting. Meanwhile, the proposed method still relies on linear structures and line lasers. In addition, the method in [27] relies on massive point-based features, however, the proposed method only needs a single straight-line object. (2) The calibration results demonstrate comparable or superior accuracy to existing state-of-the-art methods when validated against ground truth. Existing hand-eye calibration methods usually capture images of a known calibration object. Most traditional methods require a large amount of teaching work, and the filter in LVS needs to be removed. However, the proposed method directly utilizes laser image features, further improving the efficiency of calibration. Although these calibration methods work well, the proposed method still has advantages: *(1) The proposed method can achieve concurrent spatial and temporal calibration. (2) The proposed method only relies on the common straight-line weld and is teaching-free.* (3) *The* *proposed method can finish the data collection process in one minute. (4) The proposed method can be used for online diagnosis during the manufacturing process.*

### *B. Curve weld scanning experiments*

The actual calibration effect of parameters can be compared to the scanning accuracy of actual workpieces. The calibrated performance is compared with the fast calibration method [34] with and without considering time-offset through curve weld scanning experiments. Using the calibration parameters from the previous calibration experiment, we perform the curve welds scanning experiment to evaluate the spatial measurement accuracy of the proposed method thereby validating the calibration results. The curve weld and its scanning process are shown in Fig. 3 (b). We use the LVS-based robotics system to measure the curve coordinates of the feature $f_3$ . The measurement error curves are shown in Fig. 10.

Results: As shown in Table I, the proposed method is compared with existing method about the calibration time and error statistics. With a calibrated time-offset of 18 milliseconds, the RMSE and MAE reach 0.173 mm and 0.112 mm, respectively, indicating high-accuracy weld scanning. Without time-offset compensation, the RMSE and MAE increase to 1.17 mm and 1.24 mm. Accounting for time-offset reduces the RMSE by 85.3% (5.76-fold) and the MAE by 90% (10-fold). Therefore, temporal calibration plays an essential role in improving the accuracy of the weld scanning. (2) When the calibrated parameters from the proposed method are adopted, we achieve comparable or even better performance with existing state-of-the-art methods. In summary, our automatic calibration method meets the stringent accuracy requirements for dynamic welding applications.

Table IRMSE and MAE statistics.

| Method | Calibration time (second) | RMSE (mm) | MAE (mm) |
|---|---|---|---|
| Proposed method | Within 60 | 0.173 | 0.112 |
| Proposed method w/o temporal calibration | Within 60 | 1.17 | 1.24 |
| Fast calibration [34] | Within 120 | 0.163 | 0.155 |

Fig. 10. The C-type weld measurement error curve.

## V. CONCLUSION

To address the spatiotemporal calibration of LVS in robotic hand-eye systems, we propose a novel calibration method based on straight-line constraints. The proposed methodology eliminates traditional teaching procedures through an S-shaped trajectory protocol that enables programming-free path execution. The theoretical basis of this method based on straight-line constraints is described and the experimental results are given and discussed. The experimental results demonstrate that the time-offset and hand-eye parameters can be calibrated by nonlinear optimization based on straight-line constraints when the robot moves by an S-shaped path under various orientations. In addition, time-offset calibration enables data correlation without hardware support. As fundamental features prevalent in industrial environments (e.g., weld seams and structural edges), they provide computational robustness while ensuring practical implement ability. Notably, the proposed temporal calibration achieves hardware-agnostic data synchronization through pure algorithmic means, significantly reducing system integration complexity. In future work, planar constraints will be studied to improve calibration accuracy.